\documentclass[conference]{IEEEtran}
\IEEEoverridecommandlockouts
\usepackage{cite}
\usepackage{amsmath,amssymb,amsfonts}
\usepackage{algorithmic}
\usepackage{graphicx}
\usepackage{textcomp}
\usepackage{xcolor}
\usepackage{multirow}
\usepackage{caption}
\usepackage{booktabs}

\def\BibTeX{{\rm B\kern-.05em{\sc i\kern-.025em b}\kern-.08em
    T\kern-.1667em\lower.7ex\hbox{E}\kern-.125emX}}
\begin{document}

\title{You Only Speak Once to See\\
\thanks{This work was supported by National Key R\&D Program of China (No.2023YFB2603902) and Tianjin Science and Technology Program (No. 21JCZXJC00190).}
}

\author{
    \IEEEauthorblockN{
        Wenhao Yang\IEEEauthorrefmark{1}, 
        Jianguo Wei\IEEEauthorrefmark{1}, 
        Wenhuan Lu\IEEEauthorrefmark{1}, 
        Lei Li\IEEEauthorrefmark{2}\IEEEauthorrefmark{3}
    }
    \IEEEauthorblockA{
        \IEEEauthorrefmark{1}Tianjin University 
        \IEEEauthorrefmark{2}University of Washington 
        \IEEEauthorrefmark{3}University of Copenhagen \\
        \small
        yangwenhao@tju.edu.cn, lilei@di.ku.dk
    }
}

\maketitle

\begin{abstract}

Grounding objects in images using visual cues is a well-established approach in computer vision, yet the potential of audio as a modality for object recognition and grounding remains underexplored. We introduce YOSS, "You Only Speak Once to See," to leverage audio for grounding objects in visual scenes, termed Audio Grounding. By integrating pre-trained audio models with visual models using contrastive learning and multi-modal alignment, our approach captures speech commands or descriptions and maps them directly to corresponding objects within images. Experimental results indicate that audio guidance can be effectively applied to object grounding, suggesting that incorporating audio guidance may enhance the precision and robustness of current object grounding methods and improve the performance of robotic systems and computer vision applications. This finding opens new possibilities for advanced object recognition, scene understanding, and the development of more intuitive and capable robotic systems.


\end{abstract}


\section{Introduction}

Visual grounding aims to locate the most relevant object or region within an image based on a human-provided description. This task is fundamental in bridging visual perception with linguistic understanding and is essential for enhancing interaction capabilities in robotic and artificial intelligence (AI) systems \cite{mao2016generation,yu2016modeling}. Traditionally, visual grounding has been predominantly based on text and image modalities. With the advent of pre-trained large language and image models \cite{kenton2019bert,floridi2020gpt,dosovitskiy2020image,bao2021beit}, text-image models have been able to learn rich semantic information from extensive language corpora. However, the utilization of spoken language for object localization has remained largely unexplored.

Incorporating speech into visual grounding tasks presents new opportunities for advancing human-computer interaction, particularly within the domains of robotics and AI systems. Speech represents a natural and intuitive mode of communication for humans, and enabling machines to comprehend and act upon spoken instructions can significantly enhance their accessibility and usability. For example, in robotics, a user might verbally instruct a robot to "Get around the front," necessitating the robot to ground this spoken instruction by accurately identifying and locating the specified object within its visual environment. This capability would substantially benefit applications in assistive robotics, autonomous systems, and interactive AI agents, where seamless and natural interaction modalities are paramount.

Despite the considerable potential, the application of speech for object localization remains underexplored. Current research has primarily focused on text-based grounding, leaving a significant gap in leveraging auditory information for visual tasks \cite{li2022grounded,liu2023grounding,cheng2024yolo} . Addressing this gap is crucial for developing more versatile and intuitive AI systems that can interact with users through multiple modalities. Consequently, there is a pressing need for research that investigates the integration of spoken language into visual grounding frameworks, thereby expanding the capabilities and effectiveness of robotic and AI systems in real-world applications.

\begin{figure}[!t]
\begin{center}
\includegraphics[width=0.95\linewidth]{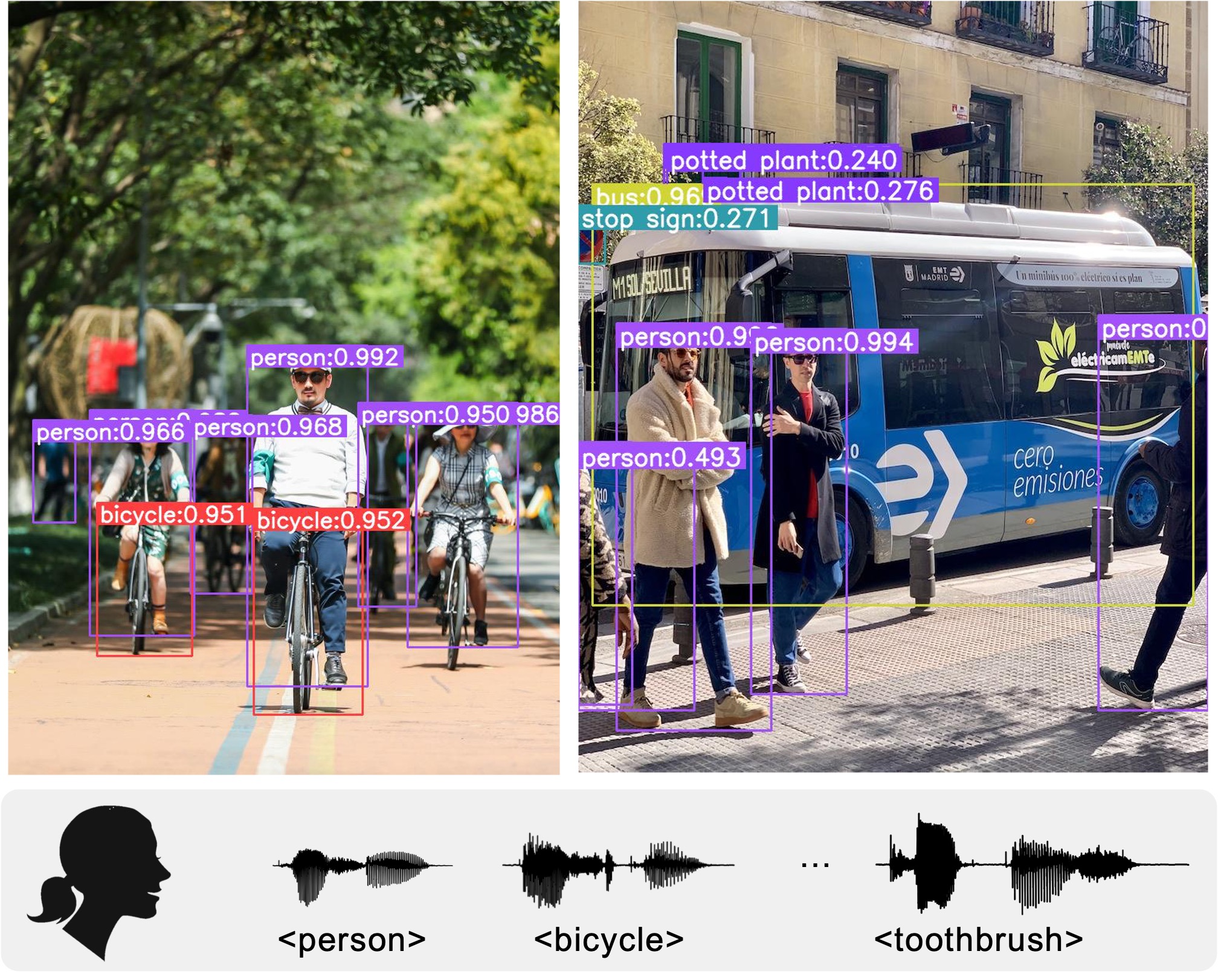}
\end{center}
\caption{Model predictions on the COCO classes with YOSS.}
\label{fig:coco_det}
\end{figure}

Our work leverages speech as a cue for visual grounding, introducing the \textbf{Audio Grounding} task. We further explore audio-based object detection adaptable to an open-vocabulary setting, as shown in Figure \ref{fig:coco_det}, advancing scene understanding and enhancing the capabilities of robotic systems. The human auditory system can process speech to extract necessary information for social interaction. In human-computer interaction, modalities beyond text, such as speech and body language, play crucial roles. Among these, audio-based dialogue interaction offers a more user-friendly and convenient means of engagement \cite{benzeghiba2007automatic,yu2016automatic}.

For language-visual grounding, leveraging paired texts and images via contrastive learning \cite{gan2020large,cheng2024open,Radford2021LearningTV,goel2022cyclip,li2024cpseg}, multi-modal models can now effectively learn text-image representations, providing a foundational application for language-visual grounding tasks. One influential model is CLIP \cite{Radford2021LearningTV}, which, due to its broad coverage of visual concepts, is semantically rich enough to be applied to downstream tasks in zero-shot settings, such as image classification \cite{abdelfattah2023cdul}, object detection \cite{guopen}, and segmentation \cite{Zhou_2023_CVPR}. With the advent of grounding tasks, open-set (open-vocabulary) object detection has become a new trend in modern computer vision. Recent works using pre-training schemes have developed numerous detector models. For example, DetCLIP \cite{yao2022detclip} and GLIP \cite{li2022grounded} present frameworks for open-vocabulary detection based on phrase grounding. Grounding DINO \cite{liu2023grounding} combines a transformer-based detector DINO \cite{zhangdino} with grounded pre-training. YOLO-World \cite{cheng2024yolo} uses pre-trained CLIP for language-image representation and employs YOLO \cite{Jocher_Ultralytics_YOLO_2023} as its detection backbone.

\begin{figure*}[!th]
\begin{center}
\includegraphics[width=0.9\linewidth]{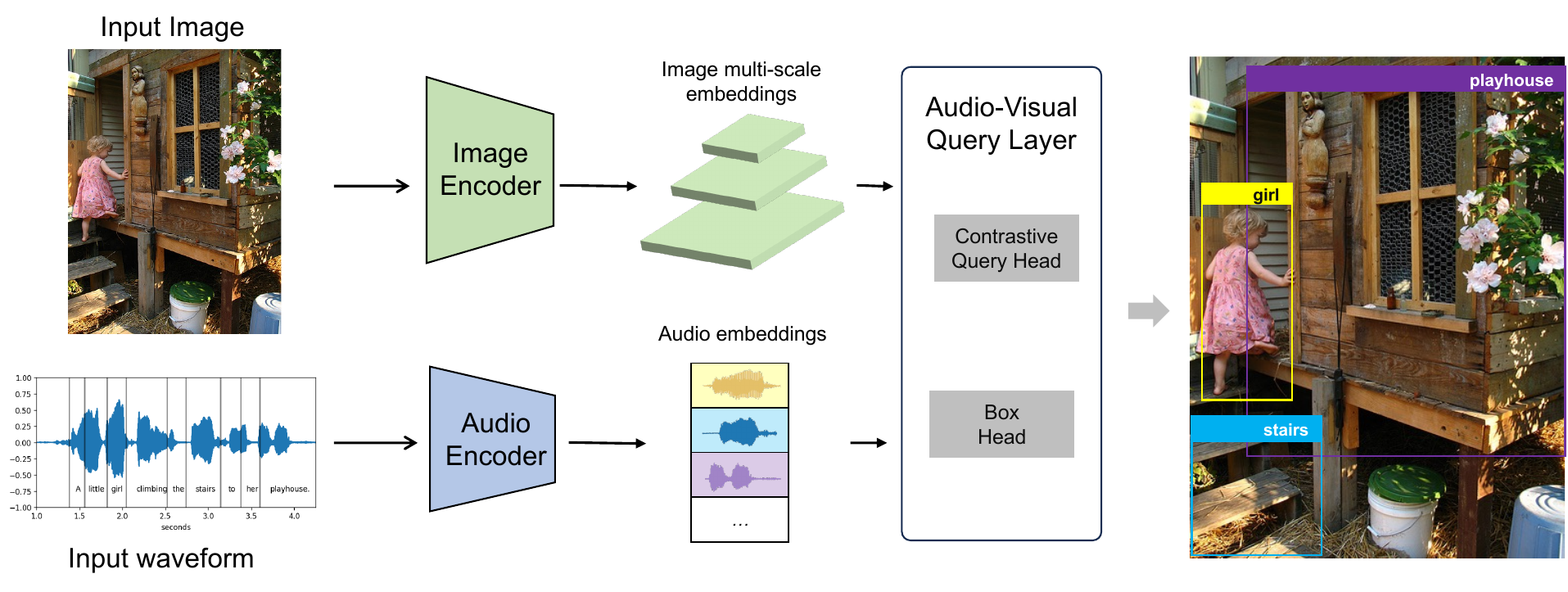}    
\end{center}
\caption{The YOSS framework for proposed Audio Grounding task.} 
\label{fig:framework}
\end{figure*}


Recent advancements in spoken language processing have significantly enhanced the ability of machines to accurately comprehend spoken input. Analogous to the CLIP model in the text-image domain, approaches like SpeechCLIP \cite{10022954} and AudioCLIP \cite{guzhov2022audioclip} perform contrastive learning across speech and text-image modalities. These methods align audio representations with visual and textual counterparts, enabling cross-modal retrieval and understanding. Consequently, unifying audio and image modalities within a single task is becoming increasingly promising and practical. In this context, our proposed framework, \textbf{YOSS} (You Only Speak Once to See), demonstrates significant potential for utilizing audio to track and understand the scene. By leveraging spoken language directly for object localization, YOSS bridges the gap between auditory and visual modalities, facilitating more intuitive human-computer interactions and advancing the capabilities of multimodal AI systems.

In summary, building upon this foundational research, we propose the Audio Grounding task in this paper. Our main contributions are:

\begin{enumerate}
\item Proposing the Audio-Image Grounding task, which uses audio prompts for open object detection;
\item Developing an Audio-Image Grounding framework that integrates multi-modal information for image and audio alignment;
\item Demonstrating the effectiveness of our framework through experiments on COCO, Flickr, and GQA datasets, and further evaluations.
\end{enumerate}

\section{Related Work}
\label{sec:related}
\subsection{Visual Grounding and Open-Vocabulary Object Detection}

Visual grounding has been extensively studied in the context of aligning textual descriptions with visual content. Early works focused on mapping phrases to image regions using attention mechanisms\cite{li2023mask,li2024cpseg,zhang2023attention}. With the advent of large-scale datasets and powerful models, approaches like CLIP \cite{Radford2021LearningTV} have demonstrated strong capabilities in learning joint text-image embeddings, enabling zero-shot transfer to various tasks such as image classification \cite{abdelfattah2023cdul}, object detection \cite{zhang2023attention}, and segmentation \cite{Zhou_2023_CVPR}. Recent models like Grounding DINO \cite{liu2023grounding} and GLIP \cite{li2022grounded} have advanced open-vocabulary object detection by integrating grounding pre-training with transformer-based architectures. These methods leverage large-scale pre-training to learn rich semantic representations, which are crucial for tasks that require understanding complex visual scenes and diverse vocabularies.

\subsection{Speech and Audio-Visual Alignment}

Despite significant progress in visual grounding using textual input, the exploration of grounding with spoken language remains comparatively underdeveloped. Several studies have investigated audio-visual alignment, focusing on tasks such as audio-visual speech recognition \cite{chung2017lip}, where models learn to recognize speech by analyzing lip movements in videos, and sound source localization \cite{arandjelovic2018objects}, which aims to identify the spatial origin of sounds within visual scenes. Cross-modal retrieval is another area where audio and visual modalities have been integrated. Works like AVLNets \cite{rouditchenko2020avlnet} have explored learning joint representations of audio and visual data to facilitate retrieval tasks across modalities. Methods such as SpeechCLIP \cite{elmanfoudi2022speechclip} and AudioCLIP \cite{guzhov2022audioclip} have extended contrastive learning frameworks to align speech and audio with text and image embeddings. By performing joint training on audio, text, and image data, these models learn shared representations that enable cross-modal retrieval and understanding.

\subsection{Self-Supervised Speech Models and Multimodal Integration}

Recent advancements in self-supervised speech models have markedly enhanced the performance of automatic speech recognition (ASR) and other downstream audio tasks. Models like wav2vec 2.0 \cite{baevski2020wav2vec}, HuBERT \cite{hsu2021hubert}, data2vec \cite{baevski2022data2vec}, and Whisper \cite{radford2022robust} learn rich representations from large-scale unlabeled audio data by predicting masked segments of the input or utilizing contrastive learning objectives. These models leverage large amounts of unannotated speech data to capture acoustic and linguistic features without the need for extensive labeled datasets. The success of these models has opened new avenues for integrating speech representations into multimodal frameworks. By combining self-supervised speech models with visual encoders, it becomes feasible to develop systems capable of understanding and grounding spoken language within images. For instance, integrating HuBERT representations with visual grounding models could enable the localization of objects in images based on spoken descriptions, enhancing applications in assistive technology and human-robot interaction. This advancement is critical for developing more natural and intuitive human-computer interactions, particularly in scenarios where voice commands are essential.

\section{Methodology}
\label{sec:meth}

In this paper, we reformulate instance text annotations as region-audio annotation pairs for Audio Grounding. Specifically, the audio segments correspond to the descriptions of objects in the audio caption dataset. Therefore, we input both the image and audio into encoders to derive the corresponding object embeddings. The image and audio features are embedded into a semantic similarity space. Subsequently, the multi-scale image features, combined with audio features, are fed into the audio-guided query selection networks. The overall implementation of the YOSS framework is illustrated in Figure~\ref{fig:framework}.

\subsection{Audio-Visual Feature Extraction}

We utilize image encoders from prior research, including CLIP. The audio encoder, specifically HuBERT, serves as an audio embedding extractor, operating in a parallel branch similar to SpeechCLIP \cite{10022954}. Initially, we pre-train the audio-image pairs using a contrastive loss to align the audio and image embeddings. Additionally, although the text encoder in CLIP could be omitted, we also attempted to explicitly align the audio and text content.

\textbf{Vision Transformer} from CLIP is employed as the pre-trained image encoder for cross-modal contrastive learning. During contrastive learning, this image encoder remains frozen and is not updated.

\textbf{HuBERT with Aggregation Branch} HuBERT \cite{hsu2021hubert} is a self-supervised learning (SSL) model designed for speech tasks, employing masked prediction pre-training. It consists of a CNN feature extractor followed by transformer layers. The Aggregation Branch includes a aggregation layer and a transformer layer with a linear projection, which maps embeddings into a shared image-audio space. The aggregation layer pools a sequence of frame-level features into fixed-dimensional embeddings using a weighted sum of the speech encoder's output. Subsequently, the high-dimensional embedding is processed through a transformer layer and a dense layer to produce a lower-dimensional utterance-level representation:
\begin{equation}\label{eq4}
 e_{a} = Transformer( WeightSum(x) )
\end{equation}
where $x$ is the hidden states for HuBERT layers and $e_a$ represents the audio embeddings.

\textbf{Contrastive Loss} is utilized for audio-image contrastive learning. After projecting the multi-modal inputs into a low-dimensional space using the image and speech encoders, the similarity relationships between these samples are learned, aligning the audio captions with their corresponding images. This process is akin to Image-Audio Retrieval \cite{10022954}. The contrastive loss can be expressed as:
\begin{equation}
\label{eq4}
L_{Con}(x_i, x_a) =  - \frac{1}{B} {\sum} log \frac{exp(sim(e_a^j, e_i^j)))}{{\sum}exp(sim(e_i^j, e_a^h))}
\end{equation}
where $j,h$ is the order of audio and image pairs,  $e_a$ and $e_i$ are the embeddings for audio and image and B is the batch size.

\begin{figure}[h]
    \begin{center}
    \begin{minipage}[b]{0.9\linewidth}
      \centering
      \centerline{\includegraphics[width=0.95\linewidth]{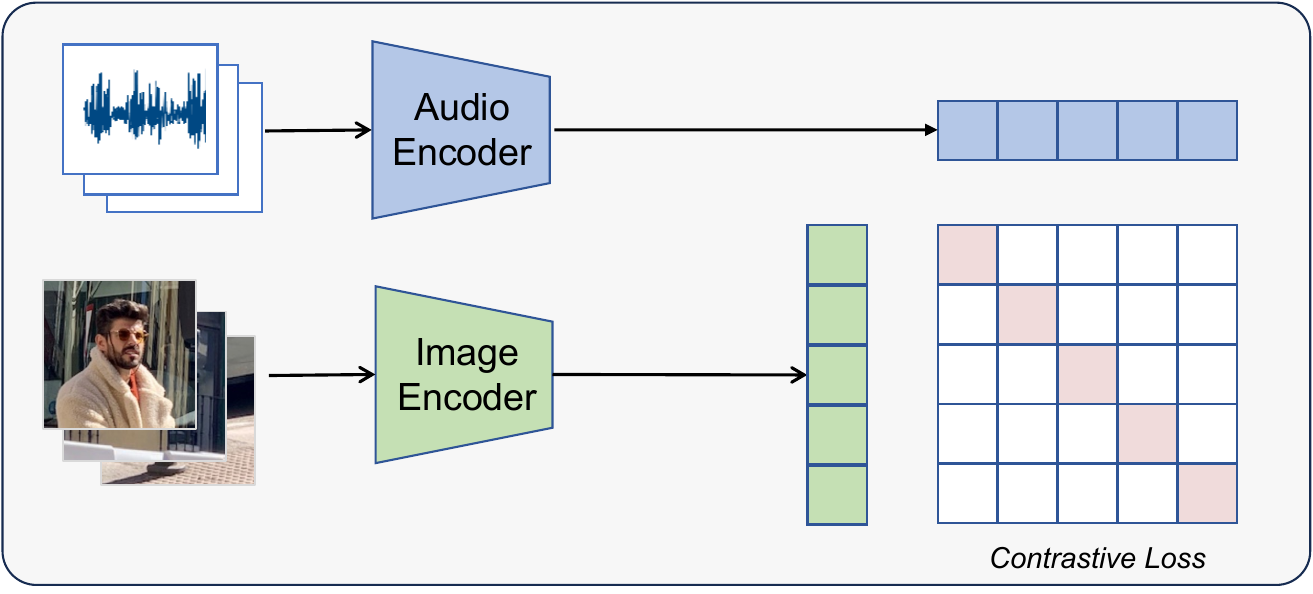}}
      \centerline{(a) Audio-Image Contrastive Learning Pre-Trained}\medskip
    \end{minipage}

    \begin{minipage}[b]{0.9\linewidth}
      \centering
      \centerline{\includegraphics[width=0.95\linewidth]{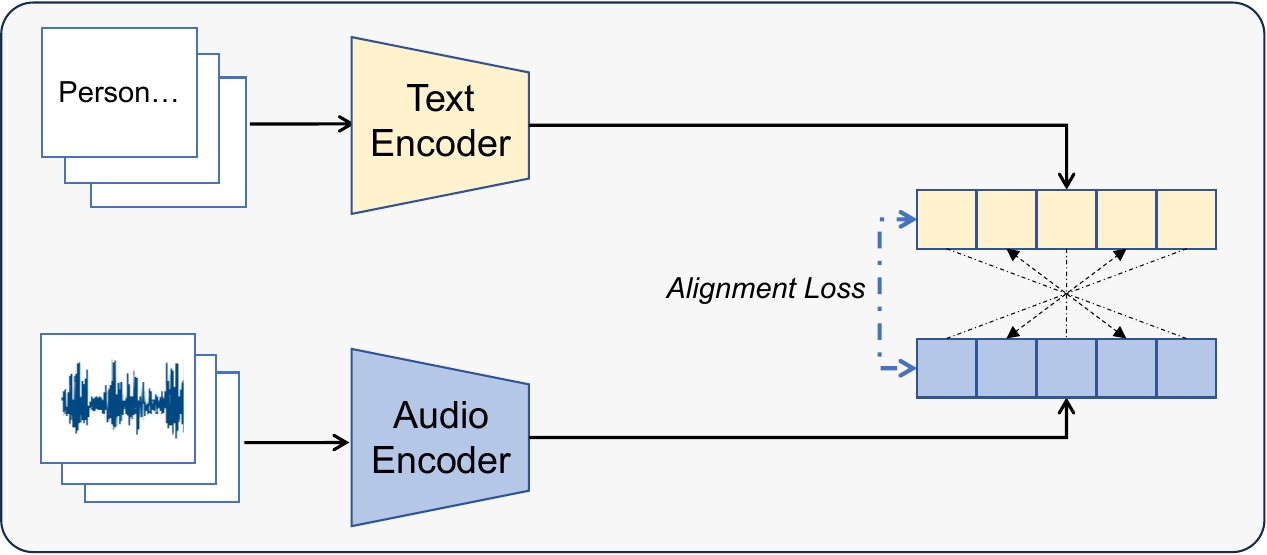}}
      \centerline{(b) Text-Audio Alignment}\medskip
    \end{minipage}
    \end{center}
    \caption{ The Contrastive and Alignemnt Learning of Audio-Visual Grounding.}
\end{figure}

\textbf{Alignment} We first align the audio and image embeddings in the CLIP embedding space. Specifically, we project multi-modal embeddings into the space of the CLIP (ViT-B/32) model using contrastive learning. Despite the contrastive learning scheme, the audio, text, and image embeddings are projected into a closely aligned common space. We propose using embedding pair alignment and correlation alignment for further refinement:
\begin{equation}
\label{eq4}
\begin{split}
 L_{Align}(e_{t}, e_{a}) = \sum L_{pair}(e_{t}, e_{a}) + \lambda \sum L_{coral}(e_t, e_a)
\end{split}
\end{equation}
where $e_t$ represents the text embeddings. It should be noted that audio-image contrastive learning can be significantly enhanced with text-modal information. Given that large text-image models are trained with vast amounts of text tokens and images, training a similar model for audio from scratch would be both time-consuming and costly.

\subsection{Audio-Visual Cross-Modal Query }

\textbf{YOLO-v8 backbone} employments YOLOv8CSPDarknet as the backbone network \cite{varghese2024yolov8}. The network outputs feature maps with diverse resolutions and dimensions. Here we use the output of the last 3 levels of feature maps for the query. 

Similar to previous Grounding work \cite{cheng2024yolo}, we use the detection head box based on NAS-FPN \cite{8954436} in YOLO-v8 for query, which is a neural architecture locating objects using pyramid features from the backbone. The multi-scale features of the image backbone and audio embeddings are combined within this process for predicting the bounding boxes and confidence scores for classes. The query consists of contrastive classification with standard CrossEntropy Loss: 
\begin{equation}
\label{eq4}
   L_{cls}(x_i, x_a) = L_{ce}( x_i, x_a ) 
\end{equation}
and localization with Distribution Focal loss \cite{li2020generalized} and IOU loss \cite{yu2016unitbox}.
\begin{equation}
\label{eq4}
   L_{loc}(x) = L_{dfl} + L_{IOU}
\end{equation}

\subsection{Unified Framework}

The overall framework for audio grounding is divided into 2 stages. First, we pre-train an audio-visual contrastive embedding model, so that audio-speech concepts are aligned with image concepts using embedding similarity learning. The pre-training loss can be expressed as: 
\begin{equation}
\label{eq4}
   L_1( x_i, x_a, x_t ) = L_{Con}( x_i, x_a ) + \eta L_{Align}( x_t, x_a ) 
\end{equation}
where $\eta$ is a constant. Then, we perform audio grounding with the YOLOv8-based cross-modal query. With audio-image pairs with annotation with audio clips and localization, the grounding query module is trained with the following loss:
\begin{equation}
\label{eq5}
   L_2( x_i, x_a ) = L_{cls}(  x_i, x_a ) + L_{loc}( x_i, x_a) 
\end{equation}

\section{Experiement}
\label{sec:format}

\subsection{Settings}

\subsubsection{Dataset}

We utilize several multi-modal image datasets to pre-train the YOSS model:

\noindent \textbf{Flickr 8K}\cite{harwath2015deep}  contains 8,000 images and 40,000 text captions with audio, which have been read aloud by 183 different speakers. 

\noindent \textbf{Flickr 30K}\cite{young2014image} contains 31,783 photographs of everyday activities, events and scenes and 158,915 text captions. 

\noindent \textbf{GQA}\cite{hudson2019gqa} is a dataset of real-world visual reasoning and compositional question answering. It consists of 133K images and 22M questions of assorted types and varying compositionality degrees, etc.  

\noindent \textbf{COCO \& SpokenCOCO }\cite{lin2014microsoft,hsu2021text} SpokenCOCO is based on the MSCOCO captioning dataset \cite{lin2014microsoft,gupta2019lvis}. This corpus was collected as a spoken version of the 605,495 captions via Amazon Mechanical Turk by displaying the text to a person and having them read it aloud. The captions were recorded by 2,352 different speakers.

The audio grounding annotation for GQA and Flickr 30k is the pseudo labeling from image-text data. The annotation from pre-trained open-vocabulary detector (GLIP) \cite{li2022grounded}, which generates pseudo boxes for each image and caption, is used for audio-grounding data synthesis. 

\begin{figure}[!t]
\begin{center}
\includegraphics[width=0.95\linewidth]{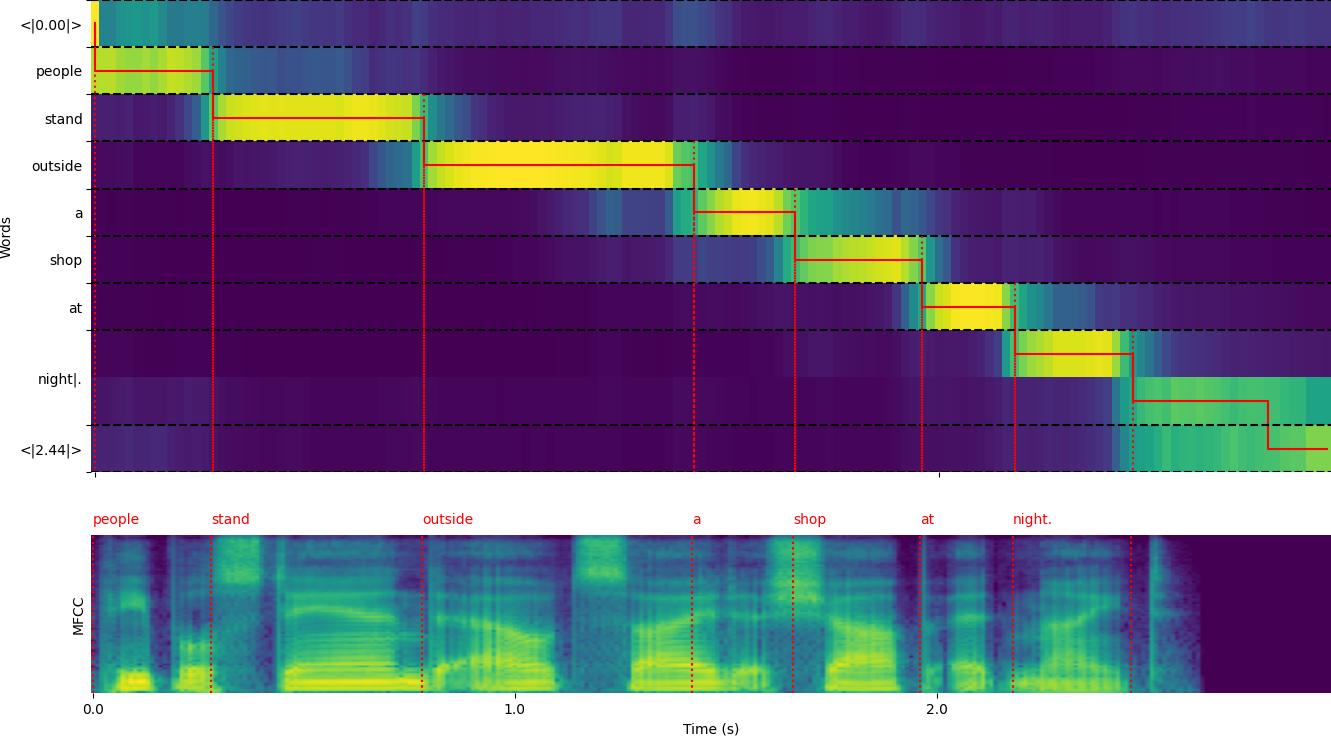}
\end{center}
\caption{Text annotation for speech utterance with timestamp whisper model.}
\label{fig:text_audio}
\end{figure}

\begin{table}[h]
\centering
\caption{Multi-Modal Datasets for the Audio Grounding task. $^\ast$ denotes synthesized audio.}
\label{tab:dataset}
\begin{tabular}{llccc}
\toprule
\multicolumn{1}{c|}{\textbf{Datasets}} & \textbf{Type} & \textbf{Images} & \textbf{Audios} & \textbf{Text} \\
\midrule
Flick 8k   & Caption & 8,091 & 40,000 & 40,461 \\
Flick 30k  & Grounding & 31,783 & 158,917 & 158,915 \\
\midrule
GQA        & Grounding & 148,854 & 71,033$^\ast$ & 22M \\
\midrule
COCO2014   & Caption & 123,287 & 605,495 & 605,495 \\
COCO2017   & Detection & 118,287 & 63,630$^\ast$ & 486,488 \\
\bottomrule
\end{tabular}
\end{table}

\subsubsection{Implement Details} We use SpeechT5 \cite{ao2022speecht5} model for Text-to-Speech synthesis in our study. To control the quality of synthesed speech, these speeches are input into the ASR model (Whisper \cite{radford2022robust}) to remove undesired ones. The details of these datasets are listed in Table~\ref{tab:dataset}. For Speech-Text Alignment, audio captions are also annotated and aligned with texts with the Whisper-based Timestamp model as in Figure~\ref{fig:text_audio}. For evaluation, the speech segments are human speech collected from the Internet.

\begin{table}[t]
\caption{Transfer of Audio Grounding results on LVIS Zero-Shot benchmarks.}
\label{tab:lvis}
\begin{center}
\begin{tabular}{ccccccc}
\toprule
\multicolumn{1}{c}{\bf Model} & \multicolumn{1}{c}{\bf Data} & \multicolumn{1}{c}{\bf Method} & \multicolumn{1}{c}{\bf AP} & \multicolumn{1}{c}{\bf AP$_r$} & \multicolumn{1}{c}{\bf AP$_c$} & \multicolumn{1}{c}{\bf AP$_f$} \\
\midrule
\multirow{3}{*}{YOSS-base} & \multirow{3}{*}{\begin{tabular}[c]{@{}l@{}}Flickr, \\ COCO\end{tabular}} & - & 6.80 & 2.30 & 5.20 & 9.00 \\
                           &                                                                            & alignment & 8.30 & 2.20 & 6.80 & 10.80 \\
                           &                                                                            & finetune & \textbf{13.60} & \textbf{4.40} & \textbf{9.50} & \textbf{18.80} \\
\midrule
\multirow{3}{*}{YOSS-large} & \multirow{3}{*}{+GQA} & - & 15.40 & 5.10 & \textbf{10.20} & \textbf{21.80} \\
                            &                       & alignment & 11.20 & 3.70 & 8.70 & 14.70 \\
                            &                       & finetune & \textbf{16.30} & \textbf{6.30} & 9.20 & 16.90 \\
\bottomrule
\end{tabular}
\end{center}
\end{table}

\begin{table}[t]
\caption{Transfer of Audio Grounding results on COCO detection benchmarks.}
\label{tab:coco}
\begin{center}
\begin{tabular}{cccccc}
\toprule
\multicolumn{1}{c}{\bf Model} & \multicolumn{1}{c}{\bf Data} & \multicolumn{1}{c}{\bf Method} & \multicolumn{1}{c}{\bf AP} & \multicolumn{1}{c}{\bf AP$_{50}$} & \multicolumn{1}{c}{\bf AP$_{75}$} \\ 
\midrule
\multirow{3}{*}{YOSS-base} & \multirow{3}{*}{\begin{tabular}[c]{@{}l@{}}Flickr, \\ COCO\end{tabular}} & - & 32.20 & 44.80 & 34.90 \\
                           &                                                                            & alignment & 33.10 & 46.20 & 35.90 \\
                           &                                                                            & finetune & \textbf{34.00} & \textbf{47.20} & \textbf{36.90} \\
\midrule
\multirow{3}{*}{YOSS-large} & \multirow{3}{*}{+GQA} & - & 37.40 & 50.80 & 40.60 \\
                            &                       & alignment & 38.40 & 52.40 & 41.70 \\
                            &                       & finetune & \textbf{39.20} & \textbf{53.30} & \textbf{42.60} \\
\bottomrule
\end{tabular}
\end{center}
\end{table}


\noindent \textbf{Training Setting} The Speech-Image contrastive learning pretraining is performed on Flickr 8k and COCO. Following the setting in SpeechCLIP \cite{10022954}, the optimizer Adam with a learning rate of 1e-4 and weight decay of 1e-6 is used. The dimension of speech embeddings is 512. With the pre-trained Speech-Image model, we finetune the Audio Grounding model with Flickr 30k, COCO and GQA for further alignment and detection. We fix the Speech Encoder and Image backbone and finetune the Audio-Visual Query layers. The AdamW optimizer is applied with a learning rate of 1e-5 and a weight decay of 0.025 for the Audio-Visual. 

The HuBERT-base model (94.7M parameters) is used in our framework. The YOSS-base model (114.9M parameters) utilizes YOLOv8-s as the backbone, and YOLOv8-m is used for YOSS-large (130.6M parameters).

\subsection{Results}

We report the transfer performance on the LVIS benchmarks and COCO object detection.

\textbf{Zero-shot Evaluation} After pre-training, we directly evaluate the proposed YOSS on the LVIS in a zero-shot manner. The images in the LVIS dataset are from the COCO dataset but are annotated with 1,203 object classes. We evaluate on LVIS minival \cite{kamath2021mdetr} and report the AP scores for our results as in Table~\ref{tab:lvis}. The maximum number of predictions is set to 1,000. YOSS models obtains acceptable result for open-vocabulary audio object detecion with limited data. However, the results lag text-based Grounding behind as spoken language understanding is relatively harder.

\textbf{Audio Object Detection} We also evaluate the proposed YOSS on the COCO2017 dataset for audio object detection. Class embeddings are generated from an audio encoder using human speech containing the corresponding class names, covering 80 object classes. We evaluate on the COCO2017 validation set and report the AP scores for our results in Table~\ref{tab:coco}. YOSS models seem to obtain better result on COCO audio object detection than open-vocabulary audio object detecion. However, the results lag text-based Grounding behind as spoken language understanding is relatively harder.

\begin{table}[h]
\centering
\caption{Alignment ablation studies in HuBERT Encoder for Image and Audio Retrieval on the Flickr 8K validation set. Recall (R@1/R@5/R@10) is reported. $*$: our implementation.}
\begin{tabular}{lcccccc}
\toprule
\multicolumn{1}{c|}{\multirow{1}{*}{\textbf{Method}}} & \multicolumn{3}{c}{Images$\rightarrow$Audio} & \multicolumn{3}{c}{Audio$\rightarrow$Image} \\
\multicolumn{1}{c|}{} & \multicolumn{1}{c}{R@1} & \multicolumn{1}{c}{R@5} & \multicolumn{1}{c}{R@10} & \multicolumn{1}{c}{R@1} & \multicolumn{1}{c}{R@5} & \multicolumn{1}{c}{R@10}\\
\midrule
MILAN \cite{sanabria2021talk}  & 49.6&79.2&87.5 &  33.2&62.7&73.9  \\
SpeechCLIP \cite{10022954}   & 41.3&73.9&84.2 & 26.7&57.1&70.0 \\
Baseline$^*$   & 46.2&76.0&86.1 & 30.1&61.8&74.5  \\
\midrule
\textbf{ +pair}  & 52.2 & 81.6 & 89.1 & 34.8 & 66.3 & 78.0   \\
\textbf{ ++coral}  & \textbf{53.0}  & \textbf{82.4} & \textbf{90.5}  & \textbf{36.1} & \textbf{67.0} &  \textbf{78.7} \\
\bottomrule
\end{tabular}
\label{tab:align}
\end{table}

In summary, it is clear that audio object detection achieves relatively acceptable results, indicating that Audio Grounding is promissing for futher exploration. Besides, alignment embeddings improve the Audio Grounding with extra information from text and further fine-tuning improves transfer results, making it possible incorping audio into multi-modal objective grounding and enhance current gounding methods.

\subsection{Ablation Studies}

\subsubsection{Alignment}

We conduct the ablation studies for the alignment of text and audio for the Audio-Visual Grounding task. The result is shown in Table~\ref{tab:align}. Within Flickr 8k val, we carried out Audio-Image Retrieval tasks. The result indicates the validation of the supervised alignment and coral alignment for cross-model learning. For recall (R@10), almost 10\% absolute increase could be observed. This also outperforms the baseline of in SpeechCLIP. 

\subsubsection{Size of Image Backbone}

The evaluation result for ablation studies of image backbone is in Table~\ref{tab:lvis} and \ref{tab:coco}. With extra parameters, YOLOv8-m contributes to a better Audio-Image Grounding model than YOLOv8-s, thus better-transferring performance of detection results.

\section{Conclusion}
\label{sec:pagestyle}

In this paper, we introduce a novel grounding task, Audio Grounding, for open-vocabulary audio object detection. We present YOSS, leveraging advancements in the grounding field. Employing audio-image contrastive learning and multi-modal alignment, we project audio and image embeddings into a shared semantic space, facilitating downstream tasks. Building on the popular detector YOLO, our audio-visual grounding framework enables immediate visual localization of spoken content.

We synthesized audio grounding datasets using a text-to-speech tool and conducted experiments on both audio-image and text-image data. Evaluations on LVIS zero-shot scenarios and COCO object detection using human speech segments demonstrate the effectiveness of our framework. Compared to text-visual grounding, our work highlights the potential of using spoken language for visual understanding. Although a performance gap exists between established text-based methods and audio grounding, it indicates the need for further research. This result also underscores the potential of incorporating the audio modality into current grounding methods to enhance their performance.

%


\bibliographystyle{IEEEtran}
\bibliography{refs}

\vspace{12pt}

\end{document}